\documentclass[letterpaper]{article} 
\usepackage{aaai25}  
\usepackage{times}  
\usepackage{helvet}  
\usepackage{courier}  
\usepackage[hyphens]{url}  
\usepackage{graphicx} 
\usepackage{natbib}  
\usepackage{caption} 
\usepackage{algorithm}
\usepackage{algorithmic}

\usepackage{graphicx}
\usepackage{subfigure}
\usepackage{rotating}
\usepackage{makecell}
\usepackage{caption}
\usepackage{adjustbox}
\usepackage{appendix}
\usepackage{booktabs} 
\usepackage{array}
\usepackage{amsmath}
\usepackage{multirow} 

\usepackage{newfloat}
\usepackage{listings}
\DeclareCaptionStyle{ruled}{labelfont=normalfont,labelsep=colon,strut=off} 
\floatstyle{ruled}
\newfloat{listing}{tb}{lst}{}
\floatname{listing}{Listing}
\title{FairTP: A Prolonged Fairness Framework for Traffic Prediction}
\author {
    Jiangnan Xia\textsuperscript{\rm 1}\equalcontrib, 
    Yu Yang\textsuperscript{\rm 2}\equalcontrib,
    Jiaxing Shen\textsuperscript{\rm 3}, 
    Senzhang Wang\textsuperscript{\rm 1}\thanks{Corresponsing author}, 
    Jiannong Cao\textsuperscript{\rm 4}
}
\affiliations {
    \textsuperscript{\rm 1}School of Computer Science and Engineering, Central South University\\ 
    \textsuperscript{\rm 2}Centre for Learning, Teaching, and Technology, The Education University of Hong Kong\\ 
    \textsuperscript{\rm 3}School of Data Science, Lingnan University\\
    \textsuperscript{\rm 4}The Department of Computing, The Hong Kong Polytechnic University\\ 
    jiangnanx129@gmail.com,
    yangyy@eduhk.hk,
    jiaxingshen@LN.edu.hk,
    szwang@csu.edu.cn,
    jiannong.cao@polyu.edu.hk
}

\begin{document}

\maketitle

\begin{abstract}
Traffic prediction is pivotal in intelligent transportation systems.
Existing works mainly focus on improving the overall accuracy, overlooking a crucial problem of whether prediction results will lead to biased decisions by transportation authorities.
In practice, the uneven deployment of traffic sensors in different urban areas produces imbalanced data, making the traffic prediction model fail in some areas and leading to unfair regional decision-making that eventually severely affects equity and quality of residents' life.
Additionally, existing fairness machine learning models fail to preserve fair traffic prediction for a prolonged time. Although they can achieve fairness at certain time points, such static fairness will be broken as the traffic conditions change. 
To fill this research gap, we investigate prolonged fair traffic prediction, introduce two novel fairness definitions tailored to dynamic traffic scenarios, and propose a prolonged fairness traffic prediction framework, namely FairTP. 
We argue that fairness in traffic scenarios changes dynamically over time and across areas.
Each traffic sensor or city area has state that alternates between ``sacrifice" and ``benefit" based on its prediction accuracy (high accuracy indicates ``benefit" state). Prolonged fairness is achieved when the overall states of sensors similar within a given period.
Accordingly, we first define region-based static fairness and sensor-based dynamic fairness. Next, we designed a state identification module in FairTP to discriminate between states of ``sacrifice" or ``benefit" to enable prolonged fairness-aware traffic predictions.
Lastly, a state-guided balanced sampling strategy is designed to select training examples to promote prediction fairness further, mitigating the performance disparities among regions with imbalanced traffic sensors.
Extensive experiments in two real-world datasets show that FairTP significantly improves prediction fairness without causing much accuracy degradation.
\end{abstract}

%
\begin{links}
    \link{Code}{https://github.com/jiangnanx129/FairTP}
\end{links}

\section{Introduction}

Traffic prediction is crucial to transportation planning, infrastructure management and optimizing resource allocation and service provision \cite{miao2022mba,miao2024unified,miao2025icde,xia2022multi,wang2024sts2anet}. 
However, existing models generally only focus on prediction accuracy, but overlook the key social impacts of traffic forecasting. In practice, achieving equitable (or fair) traffic prediction across city areas is essential, which promotes unbiased decision-making in traffic authorities and significantly improves the life quality of urban residents.

\begin{figure}[t] \centering 
    \subfigure[Traffic sensor distribution and regional performance in HK] {
            \includegraphics[scale=0.49]{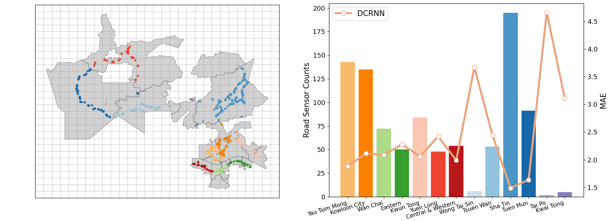}
	}
	\subfigure[Change of static fairness] { 
            \includegraphics[scale=0.35]{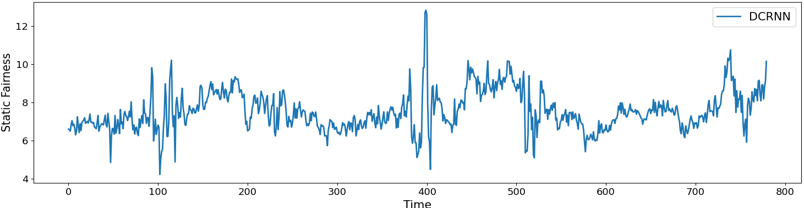}
            \label{intro_b} 
	}      
        \caption{Illustration of fairness issues in traffic prediction.} 
	\label{intro_instruction} 
\end{figure}

Actually, uneven distribution of traffic sensors deploying in a city creates data volume disparities, making the traffic prediction model fail to produce precise predictions in some underprivileged regions with fewer sensors \cite{tedjopurnomo2020survey}. 
Figure.1(a) visualizes the uneven sensor distribution in HK and the regional performance of the traffic prediction model DCRNN \cite{li2017diffusion}, where the color gradients highlight different areas. Predictive performance in regions with sparse sensors is notably lower versus the regions with more sensors deployed.
The forecast bias raises a fairness issue across city regions. Systematic underprediction of traffic in certain areas can result in insufficient transportation services, reduced ridership. It creates a negative feedback loop that amplifies existing inequities.

Existing fairness machine learning models fail to preserve fair traffic prediction for a prolonged time \cite{dong2023fairness, wan2023processing}. Although existing approaches can achieve fairness at certain time points \cite{chai2022fairness,li2024pred, guo2023towards, yang2023fara, caton2024fairness}, such static fairness may be broken as the traffic conditions change over time.
Figure.~\ref{intro_b} presents the fluctuating static fairness, clearly reflecting its temporal dynamics. This metric is calculated based on a specific definition of group fairness \cite{yan2020fairness}.
Moreover, existing methods achieve static fairness but significantly reduce accuracy in privileged areas. Enhancing fairness while achieving high prediction accuracy remains an open research problem.

To fill this research gap, the paper explores prolonged algorithmic fairness in traffic prediction, a challenging task due to several factors.
First, there lacks of clear definitions of fairness to measure and quantify equity in dynamic traffic environments.
Second, the data scarcity issue in underprivileged areas hinders accuracy improvements without negatively impacting performance in privileged regions. 
Third, achieving both shortdated static and prolonged dynamic fairness remains difficult.


To tackle these challenges, this paper introduces two novel fairness definitions for dynamic traffic environments and proposes a prolonged fairness traffic prediction framework, FairTP.
We argue that fairness in traffic prediction changes over time and across regions. 
And traffic sensors or regions have states that alternate between ``sacrifice" and ``benefit" based on their prediction accuracy. 
Prolonged fairness is achieved when overall ``sacrifice" and ``benefit" among sensors similar in a period of time. 
Accordingly, we first define region-based static fairness (RSF) and sensor-based dynamic fairness (SDF) to measure performance disparities across regions at each time point and state disparities among road sensors over a period, respectively.
Next, FairTP is proposed that consists of the following two key modules.
The state identification module discriminates ``sacrifice" and ``benefit" states, enabling prolonged fairness-aware predictions, SDF calculation, and guiding the sampling module.
Then, the state-guided balanced sampling module is designed to adjusts training examples
by increasing the sampling frequency for sensors in a "sacrifice" state. 
It improves prediction accuracy in underprivileged areas and reduces performance disparities between regions.
Lastly, both RSF and SDF are integrated into FairTP to achieve predictive fairness at shortdated static and prolonged dynamic levels.

To summarize, our main contributions are as follows.

\begin{itemize}
    \item We systematically explore prolonged fairness in traffic prediction and propose two novel fairness definitions RSF and SDF for dynamic traffic scenarios.
    \item We propose the novel FairTP framework that can seamlessly integrate with existing traffic prediction models to enhance fairness with minimal accuracy degradation.
    \item Extensive experiments on two real-world traffic datasets demonstrate that FairTP maintains high predictive accuracy while achieving both shortdated static and prolonged dynamic fairness.
\end{itemize}


\section{Related Work}
\label{section:2}

\textbf{Traffic prediction.} It has garnered significant research attention, attributable to the availability of urban data and its wide range of applications \cite{choi2022graph,xia2024multi}. 
Deep neural networks, notably Convolutional Neural Networks (CNNs) and Recurrent Neural Networks (RNNs), have gained popularity in traffic prediction due to their superior learning capabilities \cite{yao2019revisiting}. However, these models are designed for spatio-temporal grid data and are not suitable for graph-based data, which is prevalent in road networks.
Recently, there has been a rising research interest in leveraging Graph Neural Networks (GNNs) for spatio-temporal data prediction \cite{zheng2023diffuflow,yang2021time,ye2022learning}. 
Existing models have combined GNN with RNN, Temporal Convolutional Networks (TCN), or attention mechanisms to capture the complex spatial and temporal dependencies in traffic data. These models, like DCRNN \cite{li2017diffusion}, DGCRN \cite{li2023dynamic}, and DSTAGNN \cite{lan2022dstagnn}, have made advancements in capturing the dynamics of road networks.

Different from existing models, this paper pioneers a systematic study on fair traffic prediction, introduces two novel fairness definitions suitable for dynamic traffic scenarios, and develops FairTP that can seamlessly integrate with 
existing traffic prediction models and make their predictions fair with very slight accuracy degradation.

\textbf{Algorithmic Fairness.} A considerable volume of research in machine learning highlights that models can exhibit discriminatory behavior towards certain groups across different domains \cite{boratto2023counterfactual, mahapatra2023querywise, hua2023tutorial}.  
Previous fairness research has primarily focused on identifying and mitigating biases towards specific sensitive groups, such as race, in the outcomes \cite{ghani2023addressing, mehrabi2021survey}. 
Various fairness metrics, including group and individual fairness, have been proposed \cite{dong2023fairness}. 

Fairness research in transportation is in its early stages, with recent efforts exploring fairness in mobility demand
prediction \cite{du2024felight}. 
For example, Yan et al. proposed FairST, introducing two fairness metrics to promote equity across demographic groups \cite{yan2020fairness}. However, their approach relies on sensitive features like race or gender and applies regularization at time point (static).
Similarly, Zheng et al. developed SA-Net that uses socially-aware convolution operations to integrate socio-demographic and ridership data for fair demand prediction \cite{zheng2023fairness}. This method also depends on external data and static fairness regularization.
Dynamic fairness is a form of long-term fairness. It has been explored in decision-making but lacks a clear definition. Song et al. defined a dynamic fairness based on general static individual fairness \cite{song2022individual}. However, their method relies on an oracle similarity matrix created with domain knowledge or human judgment \cite{song2022guide}, making it unsuitable for traffic scenarios.


Significantly different from prior works that focus on static fairness at specific time points, we propose RSF and SDF, two novel fairness measures for dynamic traffic. They are free from sensitive attributes and can enhance predictive fairness at shortdated static and prolonged dynamic levels.


\section{Problem Statement}

Given the following: 

\textbf{Road network} $G=\{V,E\}$: A graph where $V$ represents road sensors, and $E$ indicates the connections between sensors. 
\textbf{Historical traffic data} $(X^{t-T+1}, X^{t-T+2},..., X^{t})$: Traffic observations over $T$ time steps, where $X^t=\{x_{v_i}^t | v_i \in V\}$ and $x_{v_i}^i$ is the observation of sensor $v_i$ at time $t$. 
\textbf{Region traffic conditions} $X_{Re}^t =\{x_{r_p}^t|r_p \in Re\}$: The city is divided into $m$ regions $Re = (r_1,r_2,...,r_m)$, and $x_{r_p}^t$ is the mean of all node observations in region $r_p$ at time $t$. 
\textbf{Sampled number} $N_{sam}$: The number of road sensors sampled for training.
\textbf{Dynamic time length} $T_d$: The batch length that controls the sampling frequency and enables the calculation of SDF.

The goal is to:
\textbf{Sample} $N_{sam}$ sensors from the road network $G$ every $T_d$ batches using a specified strategy. \textbf{Predict} regional traffic conditions $(X_{Re}^{t+1}, X_{Re}^{t+2},...,X_{Re}^{t+T})$ for the next $T$ time steps. 

\textbf{Objective}: Minimize prediction errors while maximizing RSF and SDF (elaborated in the methodology section).

\section{Methodology}

We begin with introducing RSF and SDF, tailored metrics for assessing shortdated static and prolonged dynamic fairness in traffic prediction scenarios, followed by an exposition of our FairTP. 

\subsection{Region-based Static Fairness}

Uneven road sensor placement causes imbalanced traffic prediction and further leads to serious fairness problems. Existing fairness models rely on sensitive features and focus only on fairness at specific time points, failing to capture the dynamic nature of traffic. Therefore, we propose RSF, a novel fairness that avoids using sensitive features and shortdated static fairness in dynamic traffic scenarios.


Different from existing group equity \cite{dong2023fairness} that relies on sensitive attributes, there is no additional sensitive information (e.g. race) in our traffic scenarios. Thus, the proposed RSF focuses on measuring the disparity in predictive performance between two regions at each time point without using sensitive information.
Areas with fewer road sensors suffer from larger prediction errors due to smaller data volume. This data imbalance in different areas of a city makes the model to perform poorly in certain areas, leading to unfair regional decisions and impacting residents' equity. To this end, we can directly reduce the performance difference by constraining the gap between regions.


\textbf{RSF}. We now formally define $RSF(r_p,r_q)$ between region $r_p$ and $r_q$ at time point $t$ as follows
\begin{equation}
    \begin{split}
        \begin{aligned}
        RSF(r_p,r_q)=\vert\mathcal{M}[\hat{y}_{r_p}^t]-\mathcal{M}[\hat{y}_{r_q}^t] \vert ,
        \end{aligned}
    \end{split}
    \label{RSF_metrics}
\end{equation}
where $\hat{y}_{r_p}^t$ represents the predicted region traffic condition for region $r_p$ at time $t$. And $\mathcal{M}[\hat{y}_{r_p}^t]$ is the mean absolute percentage error for region $r_p$ at time $t$. RSF quantifies the predictive performance disparity between two regions in the city at each time point. A smaller RSF value indicates higher fairness.


\textbf{RSF Loss}. We define the RSF loss at time $t$ as follows

\begin{equation}
    \begin{split}
        \begin{aligned}
        L_{RSF}= \frac{2}{m(m-1)}\sum_{r_p,r_q \in Re}\vert \mathcal{M}[\hat{y}_{r_p}^t]-\mathcal{M}[\hat{y}_{r_q}^t] \vert,
        \end{aligned}
    \end{split}
    \label{RSF_regular}
\end{equation}
where $m$ is the number of regions. The average of the difference in predictive performance between all regional pairings is measured by $L_{RSF}$. It is calculated at each time point. To maximize shortdated static fairness in predictions, we aim to minimize $L_{RSF}$ during training.

\subsection{Sensor-based Dynamic Fairness}

RSF measures immediate prediction errors across regions at individual time points but cannot make the prediction model preserve fair traffic prediction for a prolonged time. 
In addition, directly optimizing RSF may reduce accuracy in privileged regions with dense sensors, forcing them to compromise for underprivileged regions with fewer sensors. 
However, this does not necessarily improve the performance of underprivileged regions. As a result, while the RSF value may decrease, the performance decline in privileged regions remains unfair.


To this end, we propose SDF, a novel sensor-based metric for dynamic traffic scenarios. SDF ensures prolonged fairness by evaluating the overall state discrepancy between sensor pairs over a duration $T_d$. Different from existing individual equity methods \cite{dong2023fairness}, SDF has dynamic characteristics, requires no domain knowledge.
We argue that fairness in traffic scenarios evolves over time. To achieve fairness, the state of each traffic sensor should alternately ``sacrifice" or ``benefit" based on its prediction accuracy. And these states are identified by a dedicated state identification module. The prolonged fairness is achieved when the overall states of road sensors are similar within a defined period $T_d$.


\textbf{SDF}. For road sensors $v_i$ and $v_j$, the $SDF(v_i,v_j)$ between them is defined as follows
\begin{equation}
    \begin{split}
        \begin{aligned}
        SDF(v_i,v_j)=\vert \mathcal{D}_{T_d}[v_i] - \mathcal{D}_{T_d}[v_j] \vert,
        \end{aligned}
    \end{split}
    \label{IDF_metrics}
\end{equation}
where $d_{v_i}^{t_k}$ is the state of the road sensor $v_i$ at time point $t_k$. Its value is given by the state identification module. And $\mathcal{D}_{T_d}[v_i]= d_{v_i}^{t_1} + d_{v_i}^{t_2} + ... + d_{v_i}^{T_d}$ represents the overall state of road $v_i$ over a period of time $T_d$. 
SDF calculates the sum of state differences between all pairs of road sensors over a period. A smaller SDF value indicates higher fairness, meaning sensors are treated more consistently in $T_d$.


\textbf{SDF Loss}. We define the SDF loss during a period $T_d$ as follows
\begin{equation}
    \begin{split}
        \begin{aligned}
        L_{SDF}= \frac{2}{n(n-1)}\sum_{v_i,v_j \in V}\vert\mathcal{D}_{T_d}[v_i] - \mathcal{D}_{T_d}[v_j] \vert,
        \end{aligned}
    \end{split}
    \label{IDF_regular}
\end{equation}
where $n$ is the number of sensors used in $T_d$. $V$ is the set of road sensors that have been sampled in $T_d$.
The average of the overall state difference between all sensor pairings in the training data is measured by $L_{SDF}$. It is calculated every $T_d$ batches. To maximize prolonged dynamic fairness in predictions, we aim to minimize $L_{SDF}$ during training.

\subsection{Prolonged Fairness Traffic Prediction Framework}

\begin{figure}[!t]
    \centering
    \includegraphics[scale=0.297]{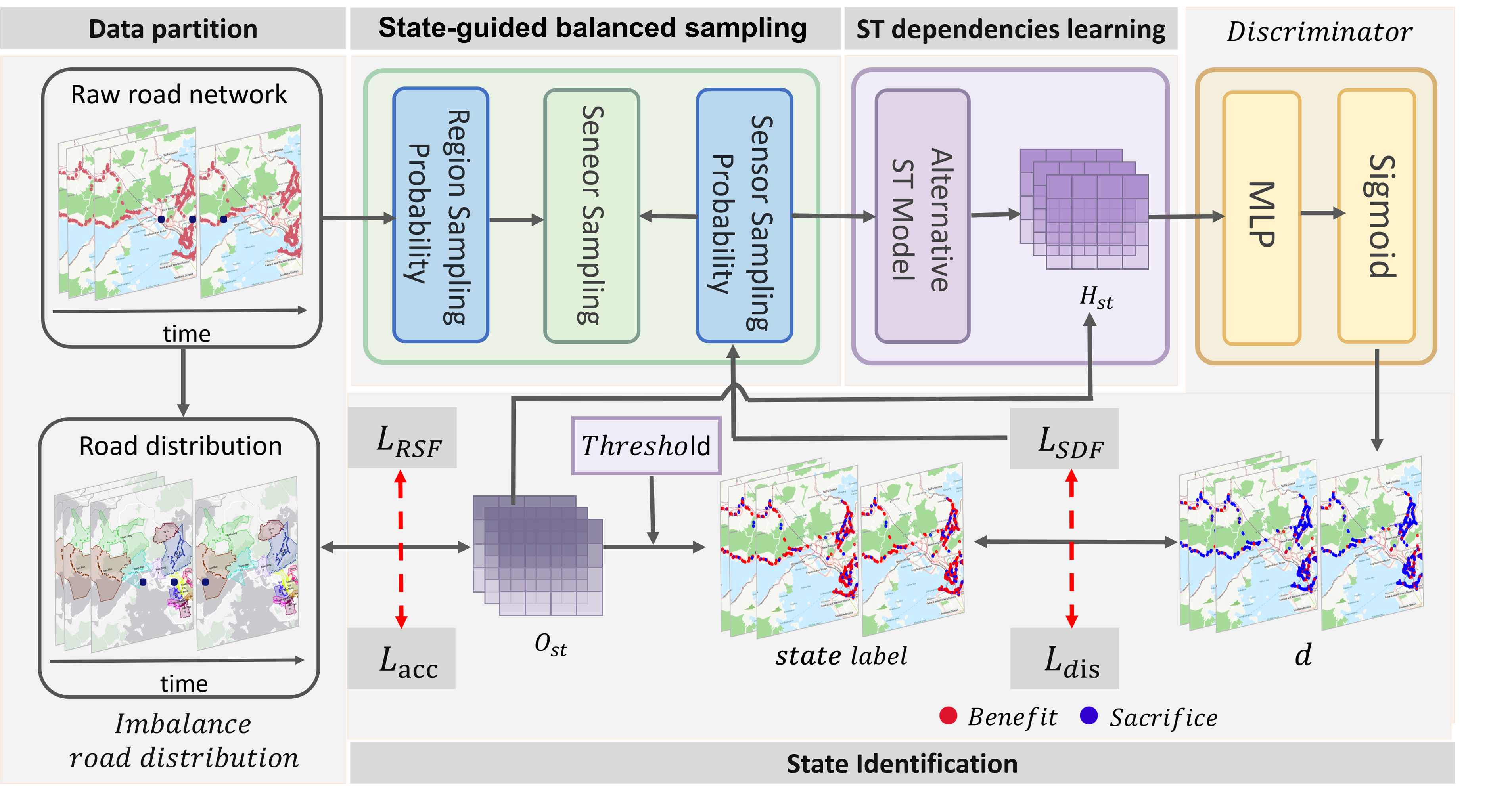}
    \caption{The framework of the proposed FairTP.}
    \label{framework}
\end{figure}

In this section, we introduce FairTP, a framework for achieving prolonged fair traffic prediction as shown in Figure~\ref{framework}. FairTP consists of three parts: (a) a state-guided balanced sampling module for sensor selection, (b) a ST dependencies learning module for learning spatio-temporal correlations, and (c) a state identification module for sensor status assessment.

\subsubsection{\textbf{State-guided balanced sampling module.}} 
Spatial imbalances in traffic data often lead to unfair predictions and biased decisions by transportation authorities.
Effective sampling of real-world traffic data is essential for building fair machine learning models. It helps counteract biases from uneven sensor deployment and reduces spatio-temporal redundancies. Moreover, it can improve models' generalization, enhance training efficiency, and balance performance between well-instrumented and under-instrumented areas.
Therefore, we propose a state-guided sampling strategy to provide balanced training data and reduce performance disparities caused by uneven sensor distribution. The strategy enhances predictions in underprivileged areas by sampling $N_{sam}$ sensors from the road network. The sampling scheme is periodically adjusted based on sensor states, giving more training opportunities to those sensors identified as ``sacrifice" (elaborated in the state identification module). This helps to improve prediction performance in these underprivileged areas.

Specifically, based on the sampled number $N_{sam}$, we start with stratified sampling to initialize the sampling. Then, after each time interval $T_d$, we calculate the sampling probabilities for all the sensors according to the results of the training feedback, and perform the next round of sampling.
A greedy algorithm is employed to adjust the sampled data every $T_d$ batches, prioritizing sensors with lower sampling probabilities until $N_{sam}$ is reached. Notably, each sampling round is influenced by the results of the previous round. Next, we describe this process in detail.

To begin the sample data collection, we use stratified sampling. This method calculates the number of road sensors in different city regions and selects a proportionate number of sensors from each region based on $N_{sam}$. It effectively reduces sampling errors and improves the representativeness and reliability of the sample. The initial sampled sensors are denoted as $Sam_0(v)$.

Second, we train the model using $Sam_0(v)$ to determine the states $d$ of the sampled sensors via the state identification module. Over a period $T_d$, these states are aggregated into an overall state $D$. And the unsampled sensors are assigned $D$=$0$.
Then, new sampling probabilities for all sensors in $V$ are calculated based on $D$. It guides the next sampling round. The process for sensor $v_i \in V$ is shown as follows
\begin{equation}
    \begin{split}
        & d_{v_i}^{t_k} =
        \begin{cases}
            d_{v_i}^{t_k} - 0.5, & v_i \in \text{Sam}_l(v), \\
           0, & v_i \in V \text{and } v_i \not\in \text{Sam}_l(v),
        \end{cases} \\
        & \mathcal{D}_{T_d}[v_i] = d_{v_i}^{t_1} +...+ d_{v_i}^{t_k}+...+d_{v_i}^{T_d} ,\\
        & P_{v_i}^l = sigmoid(\mathcal{D}_{T_d}[v_i]), v_i \in V ,
    \end{split}
    \label{DF_sample}
\end{equation}
where $d_{v_i}^{t_k}$ is the state of sensor $v_i$ at time $t_k$, determined by the state identification module. $Sam_l(v)$ represents the sensors sampled in the $l$-th round, and $\mathcal{D}_{T_d}[v_i]$ accumulates the overall state of sensor $v_i$ over the period $T_d$.
The sampling probability $P_{v_i}$ for sensor $v_i$ is calculated using a sigmoid function and is used to guide the sampling strategy for the ($l$+1)-th round.

Third, to ensure balanced sampling among different regions in the ($l$+1)-th round, we calculate the sampling probabilities for each region. This encourages an equal number of sensors to be selected from each region during the process. The formula is shown as follows
\begin{equation}
    \begin{split}
       & (P_{r_1}^l, P_{r_2}^l, ..., P_{r_m}^l) \\
         & =softmax(C_{r_1}-C_{a}, C_{r_2}-C_{a},...,C_{r_m}-C_{a}),
    \end{split}
    \label{region_sample}
\end{equation}
where $\{C_{r_p}|p=1,2,...,m\}$ denotes the number of sensors sampled in region $r_p$. $C_a = N_{sam}/m$ is the target number of sensors to be sampled in each region under balanced sampling. $P_{r_p}$ represents the sampling probability for region $r_p$. A smaller $P_{r_p}$ indicates that the number of sampling nodes in $r_p$ is farther away from the balance.


Fourth, we combine the region sampling probabilities and the sensor sampling probabilities to update the sampling probabilities for all sensors in the road network. The process is shown as follows
\begin{equation}
    \begin{split}
        \begin{aligned}
        &  P_{r_p}^l*P_{v_i}^l \xrightarrow{\text{Partition}}  P^{l+1}_{v_i}, r_p \in Re, v_i \in V,
        \end{aligned}
    \end{split}
    \label{Fusion_sample}
\end{equation}
where Partition refers to the zoning of the city, and $P^{l+1}_{v_i}$, $v_i \in V$ represents the updated sampling probability for sensor $v_i$ in the ($l+1$)-th round.

Finally, based on $P^{l+1}_{v_i}, v_i \in V$, we use a greedy algorithm to select sensors with the lowest sampling probabilities as training samples. Steps 3 and 4 are repeated, updating sampling probabilities and selecting sensors iteratively until the desired number $N_{sam}$ is reached. The new sample data $Sam_{l+1}(v)$ is then used for the next training period $T_d$. This approach ensures a representative and balanced sample across different road network regions.

\subsubsection{\textbf{ST dependencies learning module.}} 
After data sampling, we utilize a ST model to capture spatial relationships and temporal trends within the sampled data for accurate traffic forecasting. The process can be represented as follows
\begin{equation}
    \begin{split}
        \begin{aligned}
        O_{st}, H_{st} = ST(\text{Sam}(v)),
        \end{aligned}
    \end{split}
    \label{Fusion_sample}
\end{equation}
where $Sam(v)$ denotes the sampled data. $O_{st}$ represents the predicted traffic results, and $H_{st}$ is the hidden representation that contains the spatio-temporal dependencies learned by the traffic model. In this paper, the ST model is replaceable, as FairTP can be extended to any traffic prediction model.

\subsubsection{\textbf{State identification module.}} We design a state identification module that consists of a state marker and a discriminator for real-time sensor state evaluation. The discriminator is trained to infer sensor states with no ground truth during testing.

To begin with, we manually assign state labels to the sensors based on their performances. A state label of 1 indicates a ``benefit" state, and a label of 0 indicates a ``sacrifice" state. We use the output $O_{st}$ of the ST model to calculate the MAPE for each sensor. This MAPE is then compared to a predefined threshold to determine the state of the sensor. And the threshold is obtained from the previous training round.
Specifically, we argue that each sensor's prediction accuracy varies in every training round. If a sensor's MAPE is lower than the threshold, we label it as 1 (state ``benefit") for that round, indicating an improvement in performance. If the MAPE exceeds the threshold, we mark it as 0 (state ``sacrifice"), indicating a drop in performance.
Notably, the threshold is determined based on the selected ST model. We train the original ST model and record the MAPE for each round. This MAPE value is then used as the threshold when embedding the ST model into FairTP. A detailed example is provided in the appendix.

Next, we introduce a discriminator to classify sensor states at each time point. The discriminator takes the hidden representation $H_{st}$ as input and outputs a state prediction $d \in (0,1)$. We denote the discriminator as $Dis_{\theta_{dis}}: H_{st} \rightarrow d$, where $\theta_{dis}$ represents the model parameters. The discrimination loss is computed as follows
\begin{equation}
    \begin{split}
        L_{dis} = - \Big(d \log(Y)+ (1-d) \log(1-Y\Big),
    \end{split}
    \label{loss_ad}
\end{equation}
where $d$ represents the sensor states predicted by the discriminator $Dis_{\theta_{dis}}$. And $Y$ denotes the corresponding state label.
By minimizing the discrimination loss $L_{dis}$ during training, the discriminator can accurately identify the sensor states during prediction. This, in turn, guides FairTP to perform appropriate sampling, leading to fair prediction results.

\subsubsection{\textbf{Overall Objective Function.}} FairTP is trained end-to-end by minimizing a composite loss function that combines accuracy and fairness objectives
\begin{equation}
    L = L_{acc} + \lambda_1 L_{RSF} + \lambda_2 L_{SDF},
\end{equation}
where $L_{acc}$ represents the mean absolute error (MAE). $L_{RIF}$ and $L_{SDF}$ correspond to formula ~(\ref{RSF_regular}) and ~(\ref{IDF_regular}), respectively. The latter is periodically included every $T_d$ batches.
Both $L$ and $L_{dis}$ are co-train to ensure that sampling is continuously adjusted in a fair direction, ultimately yielding fair traffic predictions.

\section{Experiments}

\subsection{Experiment Setup}

\label{section:imbalance}
\textbf{Dataset.} We use two real-world datasets for regional traffic prediction: the HK and the SD datasets. The HK dataset contains six months of taxi trajectory data with 938 road sensors. The SD dataset includes data from 716 road sensors, sourced from the PeMS platform in 2019. Details are provided in the appendix.

\begin{table*}[htbp]
\small 
\renewcommand{\arraystretch}{0.97}
\setlength{\tabcolsep}{1mm} 
\centering
\begin{tabular*}{\textwidth}{@{\extracolsep{\fill}}c|ccccc|ccccc@{}}
\hline
\multirow{2}{0.07\textwidth}{} & \multicolumn{5}{c|}{HK}  &  \multicolumn{5}{c}{SD}\\
\cline{2-11}
          & \multicolumn{1}{p{0.05\textwidth}}{\centering MAE} & \multicolumn{1}{p{0.05\textwidth}}{\centering RMSE}& 
          \multicolumn{1}{p{0.05\textwidth}}{\centering MAPE} &
          \multicolumn{1}{p{0.05\textwidth}}{\centering RSF} & \multicolumn{1}{p{0.05\textwidth}|}{\centering SDF} & 
          \multicolumn{1}{p{0.05\textwidth}}{\centering MAE} & 
          \multicolumn{1}{p{0.05\textwidth}}{\centering RMSE} & 
          \multicolumn{1}{p{0.05\textwidth}}{\centering MAPE}  & 
          \multicolumn{1}{p{0.05\textwidth}}{\centering RSF} & 
          \multicolumn{1}{p{0.05\textwidth}}{\centering SDF}\\
\hline
DCRNN &\textbf{2.353}&\textbf{3.694}&\textbf{0.048}&1.779&-&28.437&44.359&0.131&2.427&- \\
FairTP-DCRNN &2.411 & 3.734 & 0.050 &\textbf{ 1.471 }&\textbf{ 0.085}&\textbf{ 21.709 }&\textbf{ 32.893 }&\textbf{ 0.113 }&\textbf{ 1.668}&\textbf{1.299} \\ 
\cline{2-11}
AGCRN &1.957&\textbf{2.974}& 0.040&1.613 &-& 19.215 & 29.067 &\textbf{ 0.078}& 1.337&-\\
FairTP-AGCRN &\textbf{1.939}& 3.243&\textbf{ 0.039}&\textbf{0.907}&\textbf{0.057}&\textbf{ 14.798}&\textbf{ 20.782}&0.082&\textbf{ 1.002}&\textbf{1.602} \\ 
\cline{2-11}
GWNET &2.189&\textbf{3.323}& 0.046& 1.688&-& 21.158& 31.117 &\textbf{0.096}& 1.206&-   \\
FairTP-GWNET  &\textbf{2.132}&3.448&\textbf{0.042}&\textbf{ 0.835}&\textbf{1.999}&\textbf{18.828}&\textbf{27.836}&0.094 &\textbf{0.945}&\textbf{6.533}\\ 
\cline{2-11}

ASTGCN &\textbf{2.005 }& 3.065 & 0.042 & 1.632 &-& 23.689 & 35.812 & 0.103 & 1.320 &-   \\
FairTP-ASTGCN  &2.110 &\textbf{ 3.264 }&\textbf{ 0.043 }&\textbf{ 0.945 }&\textbf{ 0.067 }&\textbf{ 17.965 }&\textbf{ 25.269 }&\textbf{ 0.103 }&\textbf{ 1.048 }&\textbf{ 0.614}\\ 
\cline{2-11}

DSTAGNN   &2.373 & 3.662 & 0.049 & 1.697 &-& 22.005 & 31.298 &\textbf{ 0.113 }& 1.113 &- \\
FairTP-DSTAGNN &\textbf{2.086 }&\textbf{ 3.351 }&\textbf{ 0.042 }&\textbf{ 0.973 }&\textbf{ 2.612 }&\textbf{ 18.209 }&\textbf{ 25.708 }& 0.104 &\textbf{ 0.584 }&\textbf{ 0.907} \\ 
\cline{2-11}
DGCRN &2.257 & 3.271 & 0.047 & 1.790 &-& 33.346 & 49.799 & 0.150 & 1.758 &- \\
FairTP-DGCRN  &\textbf{2.399 }&\textbf{ 3.606 }&\textbf{ 0.049 }&\textbf{ 1.241 }&\textbf{ 3.795 }&\textbf{ 21.497 }&\textbf{ 31.182 }&\textbf{ 0.109 }&\textbf{ 1.517 }&\textbf{ 4.944}\\
\cline{2-11}
D2STGNN&\textbf{1.992 }&\textbf{ 2.895 }&\textbf{ 0.042 }& 1.873 &-& 18.012 & 25.941 & 0.072 & 1.242 &- \\
FairTP-D2STGNN  &1.928 & 3.086 & 0.039 &\textbf{ 0.819 }&\textbf{ 0.739 }&\textbf{ 13.725 }&\textbf{ 19.083 }&\textbf{ 0.071 }&\textbf{ 0.736 }&\textbf{ 2.292} \\ \hline
\end{tabular*}
\caption{Performance comparison of FairTP and traffic prediction model}
\label{tab:table1}
\end{table*}

\begin{table}[h]
\small 
\centering
\renewcommand{\arraystretch}{1}
\setlength{\tabcolsep}{0.3mm}  
\begin{flushleft}
\captionsetup{justification=raggedright,singlelinecheck=false}
\caption{Performance comparison of FairTP and fairness mitigation methods.}
  \begin{tabular}{p{0.1cm}c  p{0.3cm}c p{0.7cm}c p{0.7cm}c p{0.7cm}c p{0.7cm}c p{0.7cm}c}
    \hline 
 \multirow{2}{0.02\textwidth}{} & {}  & \multicolumn{3}{c|}{HK}  &  \multicolumn{3}{c}{SD}\\
  \cline{3-8}
      {} & {}    & \multicolumn{1}{p{0.05\textwidth}}{\centering MAE} & \multicolumn{1}{p{0.05\textwidth}}{\centering RSF} & 
          \multicolumn{1}{p{0.05\textwidth}|}{\centering SDF} &
		\multicolumn{1}{p{0.05\textwidth}}{\centering MAE} & \multicolumn{1}{p{0.05\textwidth}}{\centering RSF} & 
          \multicolumn{1}{p{0.05\textwidth}}{\centering SDF} \\
\hline
\multirow{3}{*}  &
\multicolumn{1}{c|}{FairST-AGCRN}  & 1.86 &  1.87 & \multicolumn{1}{c|}{-}  & 19.33  &  1.19  &  - \\
&\multicolumn{1}{c|} {SA-Net-AGCRN}  &  1.88 &  1.65 &  \multicolumn{1}{c|}{-}  &  18.91 & 1.21 &  - \\
 &  \multicolumn{1}{c|}{FairTP-AGCRN} &  1.94 &  0.91 &  \multicolumn{1}{c|}{-}  &  14.80 & 1.00 &  - \\
  \cline{3-8}
  &  \multicolumn{1}{c|}{FairST-D2STGNN} &  2.06 &  1.86 &  \multicolumn{1}{c|}{-}  &  19.80 & 1.26 &  - \\
  &  \multicolumn{1}{c|}{SA-Net-D2STGNN} &  2.02 &  1.69 &  \multicolumn{1}{c|}{-}  &  19.26 & 1.13 &  - \\
  &  \multicolumn{1}{c|}{FairTP-D2STGNN} &  1.93 &  0.82 &  \multicolumn{1}{c|}{-}  &  13.72 & 0.74 &  - \\
 \hline
  \end{tabular}
\end{flushleft}
\label{tab:table2}
\end{table}

\textbf{Baseline.} We select several types of representative traffic prediction models as underlying ST models, including \textbf{DCRNN}\cite{li2017diffusion}, \textbf{AGCRN}\cite{bai2020adaptive}, \textbf{GWNET}\cite{wu2019graph}, \textbf{ASTGCN}\cite{guo2019attention}, \textbf{DSTAGNN}\cite{lan2022dstagnn}, \textbf{DGCRN}\cite{li2023dynamic} and \textbf{D2STGNN}\cite{shao2022decoupled}. These models help demonstrate the effectiveness and scalability of the proposed FairTP.
Additionally, we compare with fairness mitigation baselines \textbf{FairST}\cite{yan2020fairness} and \textbf{SA-Net}\cite{zheng2023fairness} are compared as fairness mitigation baselines. Details of these models are provided in the appendix.

\textbf{Implementation Details.} 
We set the sampled number $N_{sam}$ to 200 for both the SD and HK datasets. The dynamic time length $T_d$ is fixed at 3, representing 3 batches. We set the hyperparameters $\lambda_1$ and $\lambda_2$ to 0.01 and 0.1, respectively.
The proposed FairTP can be extended to various traffic prediction models. All models are implemented on the GeForce RTX 3090. Full implementation details are provided in the appendix.

\begin{figure*}[h]
        \centering
        \subfigure[AGCRN on HK]{\includegraphics[width=0.23\textwidth]{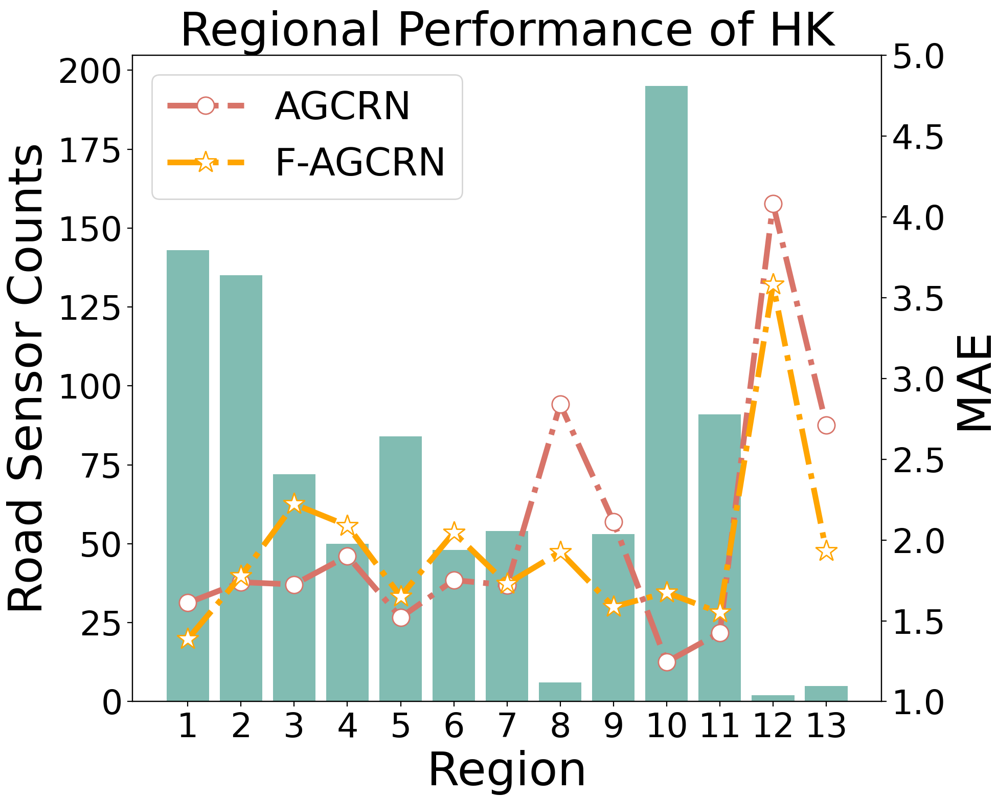}}\hspace{0.3cm}
        \subfigure[D2STGNN on HK]{\includegraphics[width=0.23\textwidth]{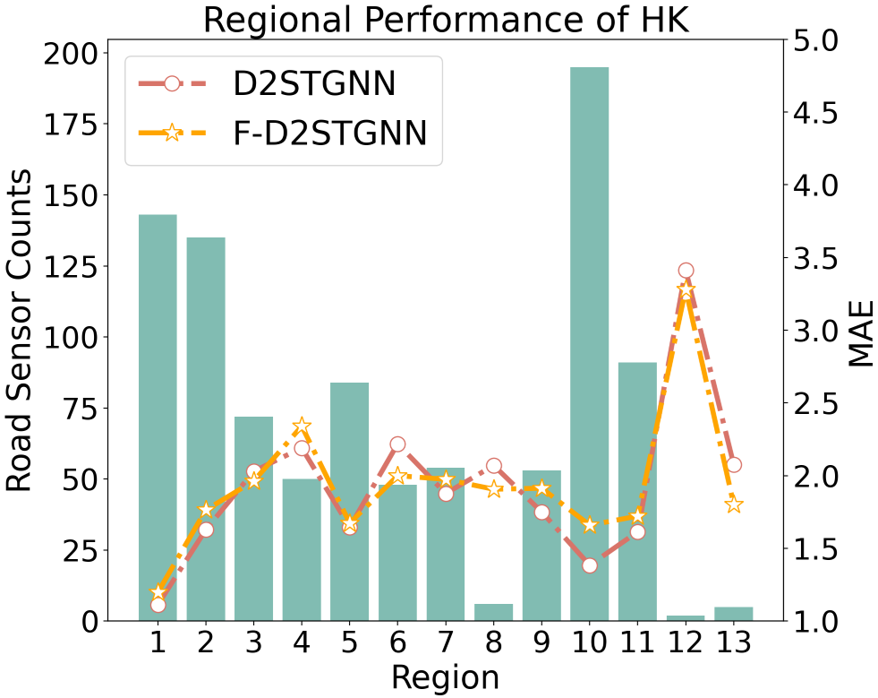}}\hspace{0.3cm}
        \subfigure[AGCRN on SD]{\includegraphics[width=0.23\textwidth]{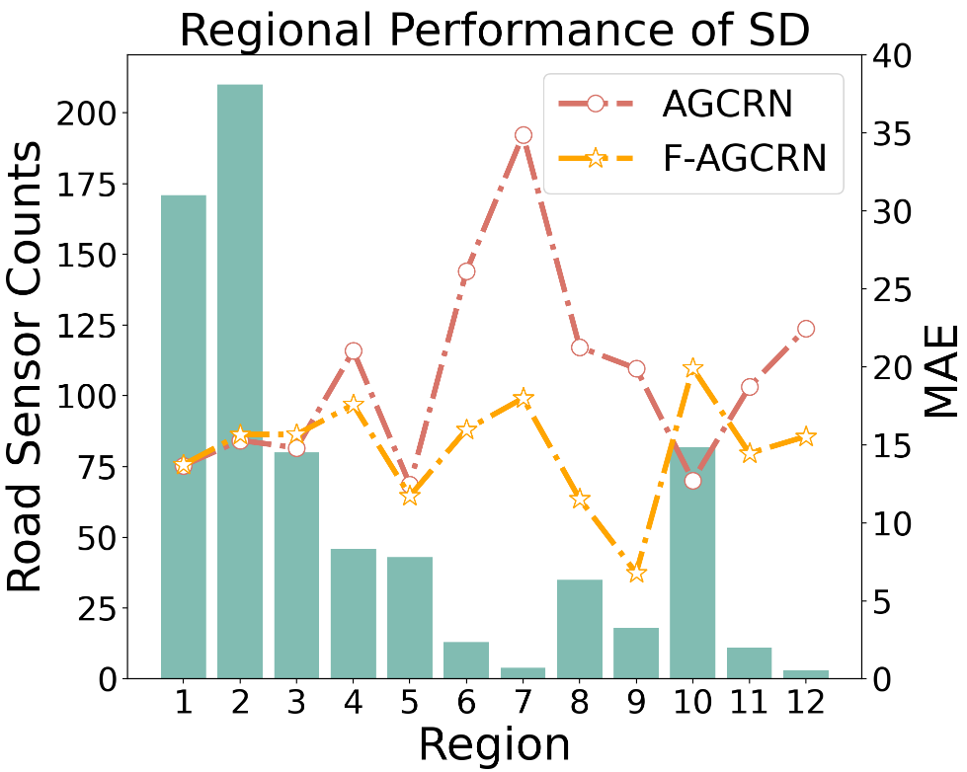}}\hspace{0.3cm}
        \subfigure[D2STGNN on SD]{\includegraphics[width=0.23\textwidth]{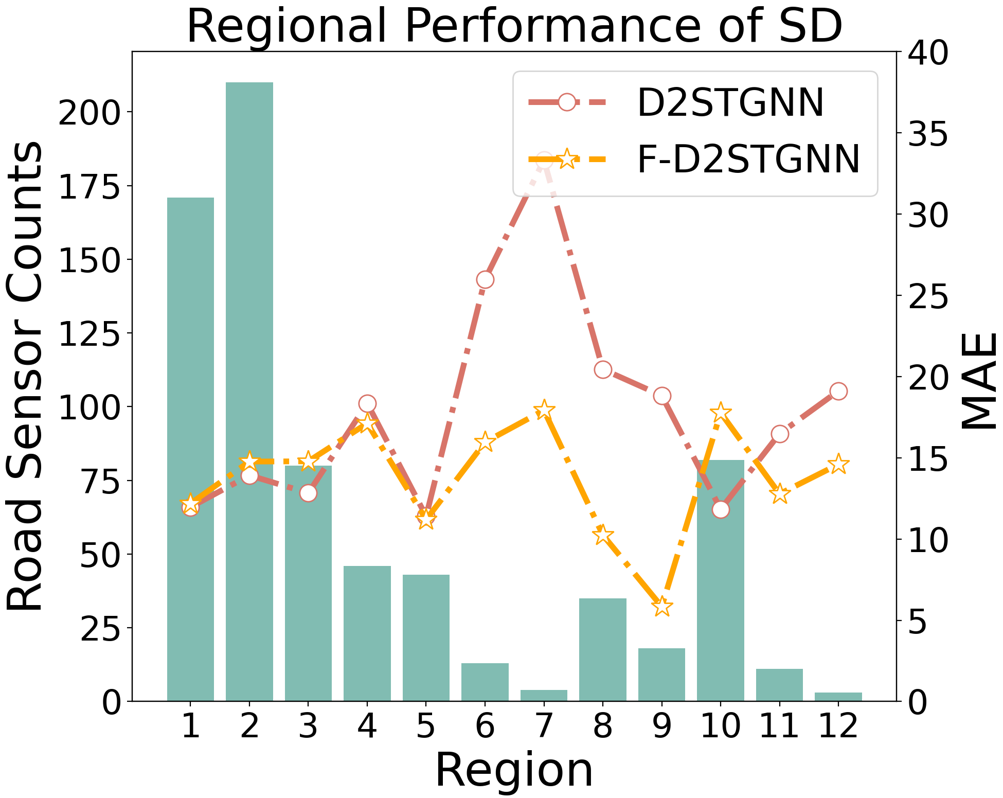}}
        \caption{Regional performance on two datasets}
        \label{fig:four-figures}
\end{figure*}

\subsection{Comparison With Baselines}

First, we extend the proposed FairTP framework to multiple baselines, denoted as $FairTP$-$baseline$. The results are shown in Table.\ref{tab:table1}, with the best performance highlighted in bold. Notably, the calculation of SDF relies on the sensor state predicted by FairTP's state identification module. Since it is absent in the baselines, they cannot produce SDF outputs.

The newly proposed FairTP ($FairTP$-$baseline$) demonstrates superior performance in most scenarios. It outperforms the corresponding baselines in terms of both fairness and accuracy.
On the HK dataset, the baselines show a reduction in MAE by 0.93\% to 13.76\%, and an improvement in RSF by 20.94\% to 128.69\%. While the MAE slightly gains (2.46\% to 5.24\%) over DGCRN, ASTGCN, and D2STGNN, the fairness improvements are substantial.
On the SD dataset, $FairTP$-$baseline$ significantly enhances the MAE performance of each baseline, with improvements ranging from 12.38\% to 55.12\%. RSF performance is enhanced by 15.89\% to 90.58\%. Notably, all $FairTP$-$baseline$ models incorporate SDF, ensuring prolonged fairness.
These results demonstrate FairTP's adaptability to various traffic prediction models, achieving strong overall performance while significantly improving shortdated static and prolonged dynamic fairness.
This improvement can be attributed to the effectiveness of data sampling, which boosts the predictive performance of underprivileged regions without significantly impacting privileged regions. Further details on the trade-off between accuracy and fairness are provided in the appendix.

Next, we compare FairTP with existing fairness mitigation methods. Due to the strict 7-page limit, we present results based on two representative baselines, AGCRN and D2STGNN, both of which perform well.

As shown in Table.2, FairTP consistently achieves the highest fairness across all cases. 
On the HK dataset, FairTP improves RSF by 82.2\% to 127.2\%, with MAE reductions ranging from 4.6\% to 6.7\%. On the SD dataset, FairTP achieves 19.0\% to 71.9\% RSF improvements and enhances MAE performance by 27.8\% to 44.3\%. FairTP achieves 19.0\% to 71.9\% RSF improvements and enhances MAE performance by 27.8\% to 44.3\%.
In summary, compared to FairST, which directly constrains predicted values, and SA-Net, which constrains MAPE values, our proposed FairTP focuses on minimizing regional MAPE differences to balance performance across areas for shortdated static fairness. Additionally, SDF is introduced to achieve prolonged dynamic fairness for road sensors.

\subsection{Regional Performance Analysis}



In this section, we provide the regional visualizations to verify the specific performance of $FairTP$-$baseline$ in the different regions.

The regional prediction performance are shown in Figure.~\ref{fig:four-figures}. In the HK dataset, $FairTP$-$baseline$ significantly improves performance in underprivileged regions such as $r_8$, $r_{12}$ and $r_{13}$, which have fewer road sensors. At the same time, the performances in the privileged regions, such as $r_{1}$, $r_{2}$ and $r_{10}$, with more sensors, remains largely unaffected.
Similarly, in the SD dataset, $FairTP$-$baseline$ shows noticeable improvements in underprivileged regions like $r_6$, $r_7$, $r_{11}$, and $r_{12}$. The performances in privileged regions like $r_{1}$ and $r_{2}$ do not observably deterioration.
We also conducted similar experiments with six other baseline models, and the results are consistent. $FairTP$-$baseline$ improves the predicted performance of the underprivileged regions on both data, while the performance of the privileged regions does not decline significantly. 
These results demonstrate that $FairTP$ indeed improves the prediction performance of underprivileged regions, leading to the better overall city-wide performance and enhancing fairness by reducing performance disparities.

\begin{table}[]
\centering
\small
 \renewcommand{\arraystretch}{1.17}
 
   \begin{flushleft}
 \captionsetup{justification=raggedright,singlelinecheck=false}
 \centering
  \caption{Performance of ablation study}
  \begin{adjustbox}{width=0.47\textwidth}
    \begin{tabular}{p{0.1cm}c p{0.3cm}c  p{0.7cm}c p{0.7cm}c p{0.7cm}c p{0.7cm}c p{0.7cm}c p{0.7cm}c}
    \hline
 \multirow{2}{0.02\textwidth}{} & {}  & \multicolumn{3}{c|}{HK}  &  \multicolumn{3}{c}{SD}\\
\cline{3-8}
      {} & {}    & \multicolumn{1}{p{0.05\textwidth}}{\centering MAE} & \multicolumn{1}{p{0.05\textwidth}}{\centering RSF} & 
          \multicolumn{1}{p{0.05\textwidth}|}{\centering SDF} &
		\multicolumn{1}{p{0.05\textwidth}}{\centering MAE} & \multicolumn{1}{p{0.05\textwidth}}{\centering RSF} & 
          \multicolumn{1}{p{0.05\textwidth}}{\centering SDF} \\
\hline 
\multirow{4}{*}{\rotatebox{90}{ \makecell{ \scriptsize FairTP-AGCRN}} }  &
\multicolumn{1}{c|}{noS}  & 2.02 &  1.01 & \multicolumn{1}{c|}{0.18}  & 14.81  &  1.12  &  2.35 \\
&\multicolumn{1}{c|} {noD}  &  1.99 &  0.88 &  \multicolumn{1}{c|}{188.20}  &  15.50 & 1.05 &  344.12 \\
 &  \multicolumn{1}{c|}{noAS} &  1.78 & 1.07 &  \multicolumn{1}{c|}{0.65}   & 12.40 & 1.23 &  2.18  \\
  & \multicolumn{1}{c|}{ALL} & 1.94 & 0.91 &  \multicolumn{1}{c|}{0.06}   & 14.80  &  1.00 &  1.60  \\ \hline

  \multirow{4}{*}{\rotatebox{90}{ \makecell{ \tiny FairTP-G2STGNN}}}  &
\multicolumn{1}{c|}{noS}  &  1.94 & 0.95 & \multicolumn{1}{c|}{0.83}   & 13.80 &  0.80 & 2.32   \\
    &  \multicolumn{1}{c|}{noD} &  2.14 & 0.83 & \multicolumn{1}{c|}{181.89} &  14.00 & 0.74 &  83.30  \\              
    &  \multicolumn{1}{c|}{noAS} &  1.73 &  1.04 & \multicolumn{1}{c|}{1.38}   &  11.42 &  0.98 &  2.47  \\
 &  \multicolumn{1}{c|}{ALL}  &1.93 & 0.82 &  \multicolumn{1}{c|}{0.74}  &  13.72 &  0.74 & 2.29\\
\hline
  \end{tabular}
       \end{adjustbox}
         \end{flushleft}
  \label{tab:table3}
\end{table}

\begin{figure}[!t] \centering
    \subfigure[FairTP-AGCRN on HK] { 
            \includegraphics[scale=0.152]
            {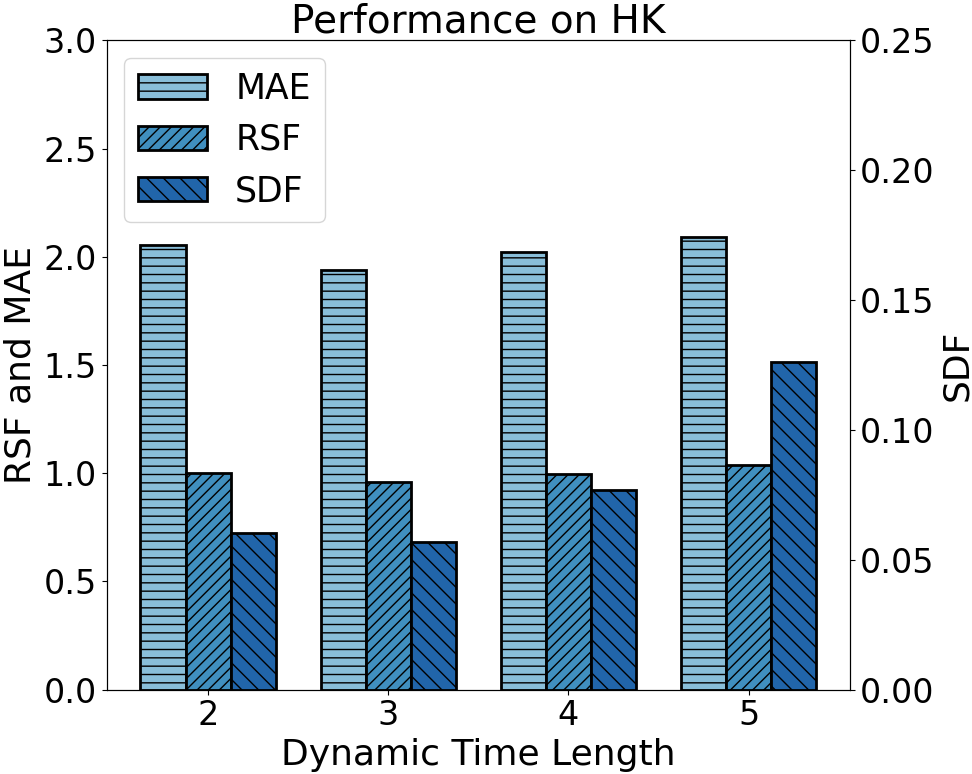}
            \label{HK_cfT} 
	} \hspace{0.0cm}
	\subfigure[FairTP-AGCRN on SD] {
            \includegraphics[scale=0.150]
            {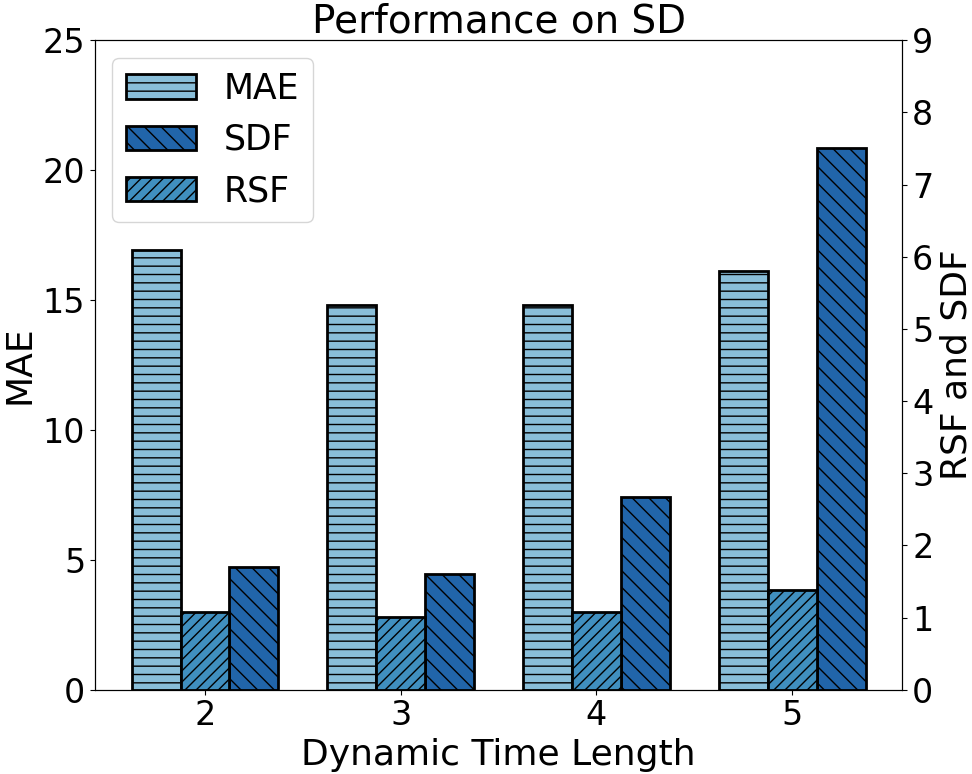}
            \label{SD_cfT} 
	}      
	\caption{Effect of dynamic time length $T_d$.}
	\label{fig5} 
 
\end{figure}

\begin{figure}[!t] \centering
    \subfigure[FairTP-AGCRN on HK] { 
            \includegraphics[scale=0.154]
            {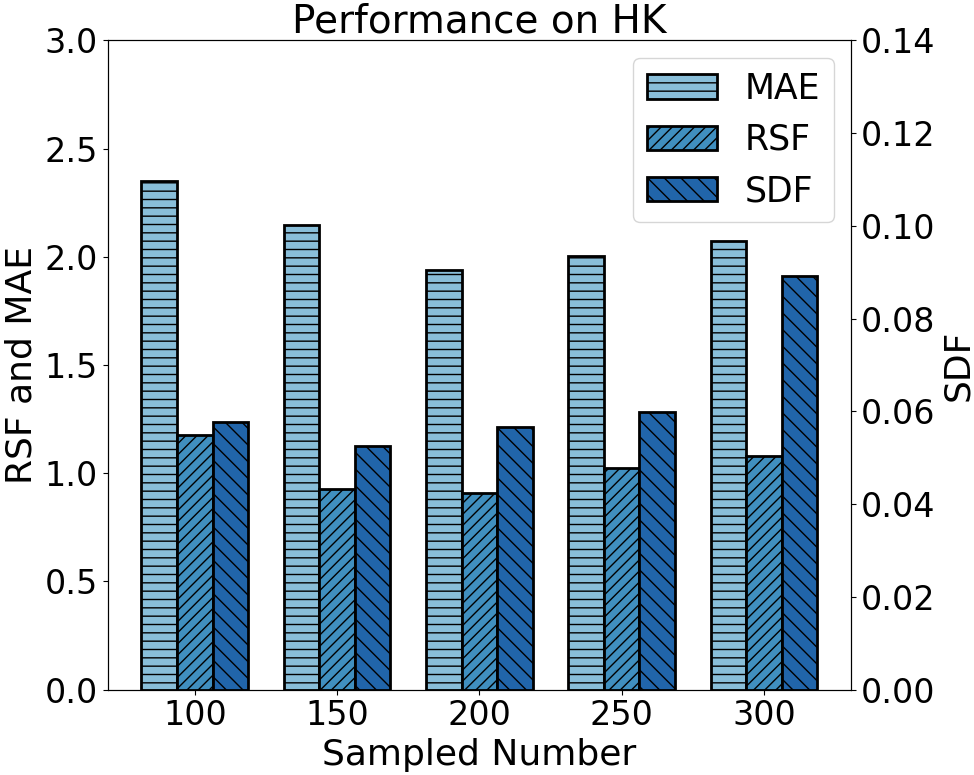}
            \label{HK_cfN} 
	} \hspace{0.0cm}
	\subfigure[FairTP-AGCRN on SD] {
            \includegraphics[scale=0.152]
            {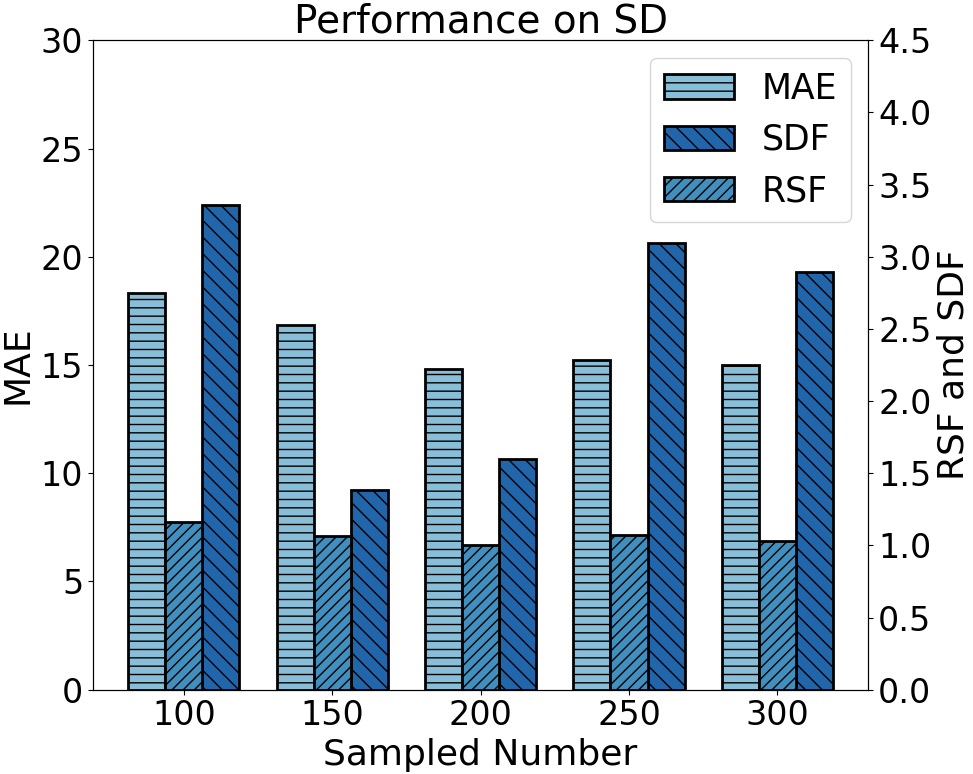}
            \label{SD_cfN} 
	}      
	\caption{Effect of sampled number $N_{sam}$.}
	\label{fig6} 
\end{figure}

\subsection{Ablation Study}

We further conduct an ablation study to evaluate the contribution of each component in FairTP to the performance gain. We deactivate different components and form the following variants. \textbf{noS} removes the $L_{RSF}$ which does not consider static fairness constraints, \textbf{noD} removes the $L_{SDF}$ which does not consider constraints on fairness at the prolonged dynamic level, and \textbf{noSA} removes the state-guided sampling module but uses the fixed stratified sampling.

The results of the ablation study are shown in Table.3. 
One can see that the $L_{RSF}$, $L_{SDF}$ and the sampling module are all useful for FairTP as removing any one of them increases the prediction error or decreases the RSF or DSF.
Among these, the $L_{SDF}$ appears to have the most significant impact. When it is removed, the SDF decreases remarkably, demonstrating the importance of the prolonged fairness. Removing the $L_{RSF}$ results in a decline in both RSF and SDF, highlighting the importance of static fairness constraints.
When using fixed stratified sampling, the performance of the model improves. It may contribute to the adaptive matrix in the baseline that is able to learn more ST information based on the fixed road sensors. 
The improvement in RSF is slightly noticeable, likely because the predictive performance of most privileged regions is already quite close, but the SDF performance significantly decreases.

\subsection{Parameter Analysis}


We investigate the effects of dynamic time length $T_d$ and sample size $N_{sam}$ on the performance of FairTP-AGCRN. The results are shown in Figure.\ref{fig5} and Figure.\ref{fig6}.
For $T_d$, we vary it from 2 to 5 and find that the best performance, in terms of RSF and SDF, occurred at $T_d = 3$. Performance declined when $T_d$ increased beyond 3, likely due to the greater fluctuations in sensor states.
Similarly, we tune $N_{sam}$ from 100 to 300. A small number of sensors result in insufficient data representation, while excessive sensors introduce noise. The optimal number of sensors for balanced performance is $N_{sam} = 200$
\subsection{Case Study}

\begin{figure}[!t]
    \centering
    \includegraphics[scale=0.27]{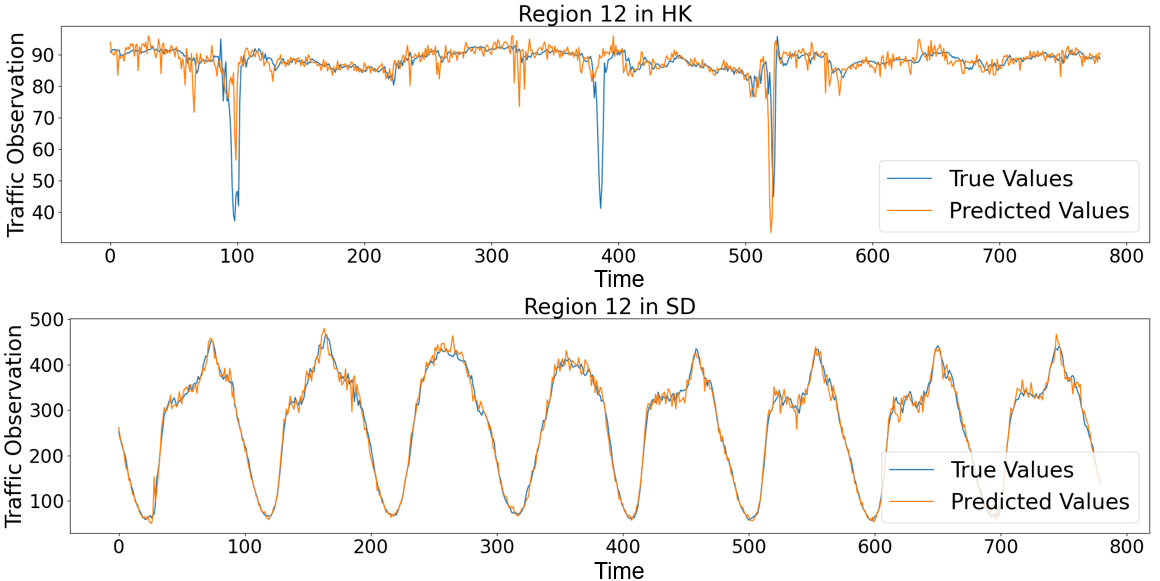}
    \caption{Case study.}
    \label{fig7}
\end{figure}


Comparative visualizations in Figure \ref{fig7} illustrate the effectiveness of FairTP by comparing its predicted traffic flow to the ground truth. The case study is conducted on underprivileged regions ($r_{12}$) with minimal sensor coverage in both HK and SD, highlights the ability of FairTP-AGCRN to accurately capture the real traffic patterns, including sharp fluctuations. The close alignment between the orange predicted curves and the blue ground truth curves demonstrates the model's high predictive accuracy and its success in improving fairness at both short-term static and long-term dynamic levels.

\subsection{Conclusion}

In this paper, we investigate prolonged fair traffic prediction for the first time. Two novel fairness definitions, RSF and SDF, are proposed for dynamic traffic prediction. Moreover, we innovatively devise FairTP. It is a prolonged fairness traffic prediction framework that can easily integrated into existing traffic models to enhance their short-term and prolonged prediction fairness with minimal impact on accuracy. 
Extensive experiments on two real-world datasets verify that FairTP keeps or even enhances prediction accuracy while significantly improving predictive fairness.

\section*{Acknowledgments}
This research was funded by the National Science Foundation of China (No.62172443), Hunan Provincial Natural Science Foundation of China (No.2022JJ30053), the Research Institute for Artificial Intelligence of Things (RIAIoT) at PolyU, the Hong Kong Research Grants Council (RGC) under the Theme-based Research Scheme with Grant No.T43-513/23-N and T41-603/20-R, Lingnan University (LU) (DB23A4) and Lam Woo Research Fund at LU (871236).

\bibliography{aaai25}

\appendix

\section{A. Terminologies}
We define some terminologies to state the studied problem.

\textbf{Road network} is represented as $G=\{V,E\}$, which reflects the spatial correlation among roads. $v_i\in V$ represents the road sensor in the city. $e_{ij}\in E$ refers to binary-valued connectivities if $v_i$ and $v_j$ are connected with each other.

\textbf{Traffic sequence data} We use $x_i^t$ represent the traffic observation of node $v_i$ at time $t$, and the observations in $T$ time points form a time series $(x_i^1,...,x_i^t,...,x_i^T)$. The traffic observations of all roads in $G$ in $T$ time points form the traffic sequence data, which can be denoted as $(X^1,...,X^t,...,X^T)$.

\textbf{Region traffic conditions}. We divide the whole city into regions like $Re = (r_1,r_2,...,r_m)$ according to certain rules, where $m$ is the number of regions. Different regions have different numbers of road nodes, we use the mean of traffic observations of all nodes at time $t$ in the $p$-th region as the corresponding region traffic conditions $x_{r_p}^t$. And $X_{Re}^t =(x_{r_1}^t,x_{r_2}^t, ..., x_{r_m}^t)$ is the region traffic conditions in city at time point $t$.

\textbf{Sampled number} represents as $N_{sam}$. It is the total number of road sensors contained in sampled training data.

\textbf{Dynamic time length} represents as $T_d$, which means the length of the batch, is used to control the frequency of sampling and calculate the SDF. 

\section{B. Stratified Sampling}
In this section, we provide a detailed description of the stratified sampling for the data in this paper.

To initiate the sample data collection, we use stratified sampling.
This calculates the number of road sensors in different areas of the city and selects a proportionate number of sensors from each region as training samples based on $N_{sam}$. It could effectively reduce sampling errors and improve the representativeness and reliability of the sample. And the initial sampled sensors are represented as $Sam_0(v)$.   

\section{C. Specific Example of State Identification Module}

In this section, a detailed description of the process in the state identification module is provided.

We argue that each sensor has different prediction accuracy in every round of training. If the prediction accuracy of a sensor is higher than a preset threshold, its performance improves in this round of training with the selected samples. We annotate it as a ``beneficial" state in this round of training and mark it as 1. Otherwise, we mark it as 0, denoting the state of performance drops. 

Taking DCRNN as an example, by running DCRNN, the MAPE of each round of training is recorded as the threshold value, and then we apply FairTP to DCRNN and get Fair-DCRNN. In each round of training of Fair-DCRNN, the states of sensors are obtained by comparing MAPE of FairTP-DCRNN to the threshold values.

\section{D. Experiment Setup Detail}
\label{Appendix-B}

\subsection{D.1 Dataset} \label{B-1}

We provide detailed descriptions of the datasets adopted in experiments.

\begin{figure}[!t] \centering
    \subfigure[Partition in HK] { 
            \includegraphics[scale=0.155]{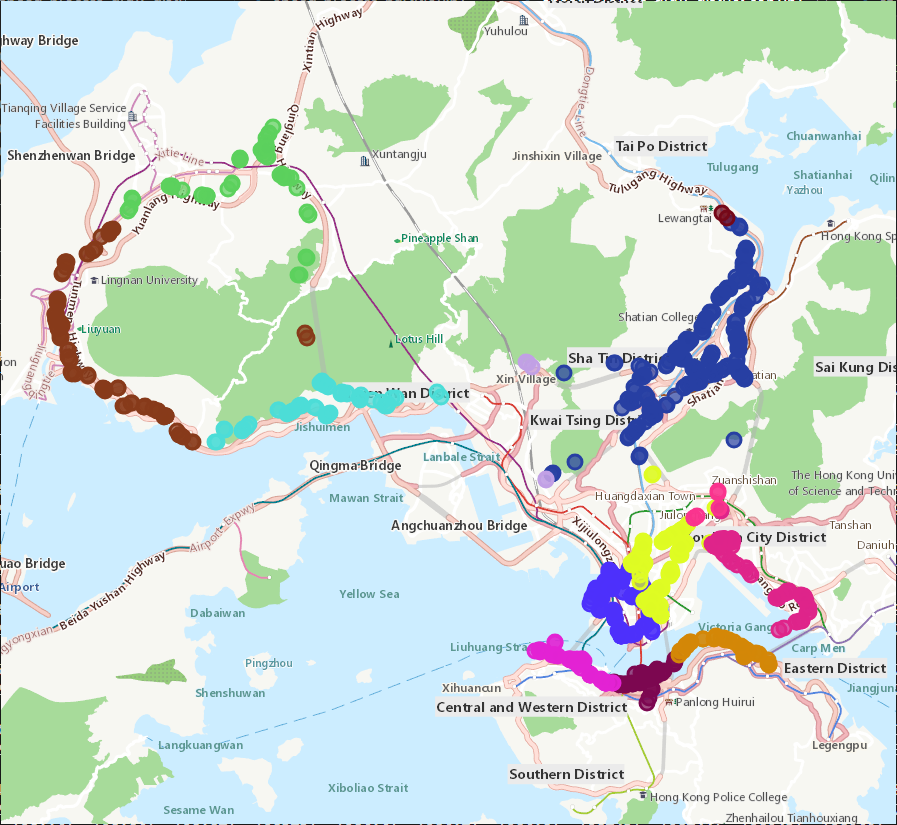}
            \label{HK} 
	} \hspace{0.2cm}
	\subfigure[Partition in SD] {
            \includegraphics[scale=0.158]{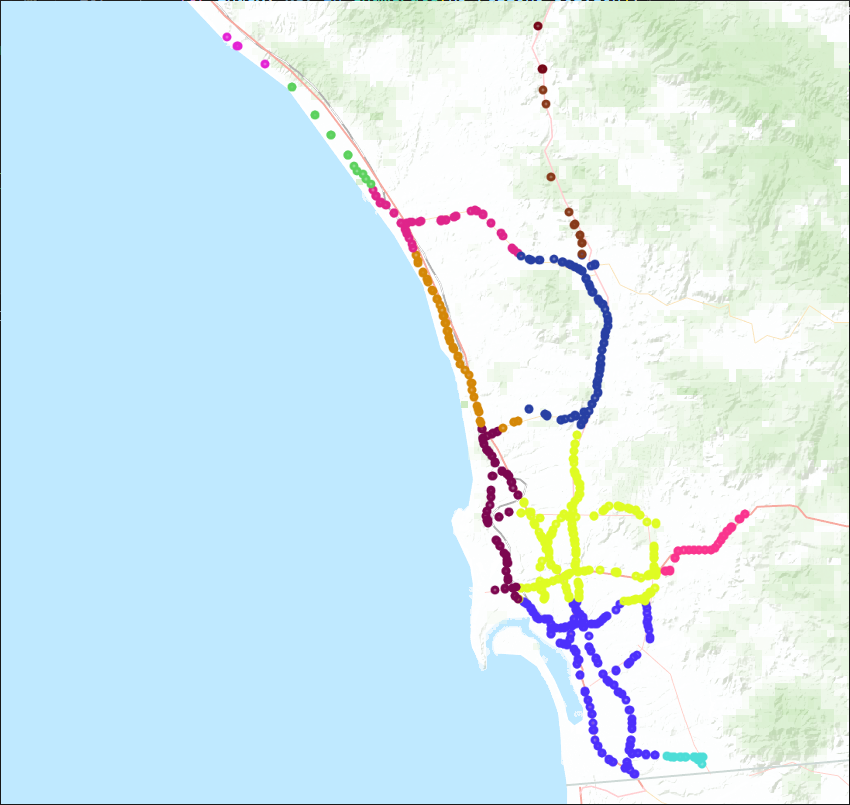}
            \label{SD} 
	}      
	\caption{Illustration of datasets partition}
	\label{into1} 
\end{figure}

\textbf{HK Didi dataset} spans from October 1, 2020 to March 31, 2021, contains trajectory data of Didi taxis over 6 months, covering 938 road sensors in HK, and traffic speed data are served as a feature. 
\textbf{SD dataset} is collected from the PeMS platform, comprises 716 road sensors for San Diego County in 2019, and we use sensor readings as traffic features.
The partitions are shown in Fig.~\ref{into1}, with different colored dots representing road sensors in different areas. HK is divided into 13 regions and SD is divided into 12 regions. 
We chronologically split the data into train, validation, and test sets, with a ratio of 6:2:2 for all datasets.

\subsection{D.2 Baseline} \label{B-2}
We elaborate baselines utilized in experiments, which contain general traffic prediction models and fairness enhanced models.
\begin{itemize}
    \item \textbf{DCRNN} \cite{li2017diffusion} used a novel diffusion convolution that works alongside GRU.
    \item \textbf{AGCRN} \cite{bai2020adaptive} used adaptive adjacency matrix and RNN to predict traffic flow. 
    \item \textbf{GWNET} \cite{wu2019graph} developed a novel adaptive dependency matrix and a stacked dilated 1D convolution component to capture ST correlations.
    \item \textbf{ASTGCN} \cite{guo2019attention} utilized ST attention mechanism to capture the dynamic ST correlations in traffic data.
    \item \textbf{STGODE} \cite{fang2021spatial} leveraged neural ordinary differential equations to effectively model the continuous changes of traffic signals. 
    \item \textbf{DSTAGNN} \cite{lan2022dstagnn} proposed a dynamic ST aware graph to learn dynamic spatial relevance among nodes.
    \item \textbf{DGCRN} \cite{li2023dynamic} generated a dynamic matrix that combines with the original road network matrix to capture more spatial information.
    \item \textbf{D2STGNN} \cite{shao2022decoupled} utilized a Decoupled Spatial-Temporal Framework (DSTF) that separates the diffusion and inherent traffic information and learned dynamic characteristics of traffic networks.
    \item \textbf{FairST} \cite{yan2020fairness} is a fairness-aware model designed for mobility prediction. It fused several convolutional branches and incorporated fairness metrics as regularization to enhance equity across demographic groups.
    \item \textbf{SA-Net} \cite{zheng2023fairness} integrated social demographics and ridership information, and introduced a bias-mitigation regularization for fair ride-hailing demand forecasting.

\end{itemize}

\subsection{D.3 Implementation Details} \label{B-3}

We obtained baseline codes from their GitHub repositories, ensuring thorough cleanup and integration for ease of comparison and reproducibility. For baselines lacking code, corresponding models were developed according to their paper specifications. 
Since FairST and SA-Net target image data, we only utilize their fairness regularization, integrating it with existing graph models for forecasting to match our graph data.
Experiments adhered to the training configurations recommended in their original sources, which were conducted on an Nvidia GeForce RTX 3090 GPU with Python 3.8, CUDA 11.1, and PyTorch 1.8.2.
For all baselines without FairTP, we use all sensors for training. 

\section{E. Trade-off between accuracy and fairness} 
\label{Appendix-C}


We evaluate all methods and compare the trade-off between predictive performance and fairness on two datasets. The results are shown in Table.5. 

One can see that the proposed $FairTP$-$baseline$ can improve the fairness in prediction while ensuring or even improving the prediction performance. In HK dataset, the improvement of MAE fluctuates from -1.14\%-13.03\%, while the improvement of the fairness metric RSF can reach 27.84\% to 94.00\%. For SD dataset, MAE performance on all methods improved by 4.86\% to 50.28\%, while RSF improved by 42.16\% to 89.01\%. This validates that our methods can improve fairness while minimizing the performance drops. 
In addition, by introducing SDF, we can further achieve fairness at prolonged dynamic level.

\begin{table}[!]
\centering
 \renewcommand{\arraystretch}{1.17}
   \begin{flushleft}
\captionsetup{justification=raggedright,singlelinecheck=false}
  \caption{Trade-off between predictive performance and fairness}
  \begin{adjustbox}{width=0.47\textwidth}
    \begin{tabular}{p{0.1cm}c  p{0.7cm}c p{0.7cm}c p{0.7cm}c p{0.7cm}c p{0.7cm}c p{0.7cm}c}
    \hline 
 \multirow{2}{2\textwidth}{} & \multirow{2}{*}{} & \multicolumn{3}{c|}{HK}  &  \multicolumn{3}{c}{SD}\\
\cline{3-8}
      {} & {}    & \multicolumn{1}{p{0.07\textwidth}}{\centering MAE} & \multicolumn{1}{p{0.05\textwidth}}{\centering RMSE} & 
          \multicolumn{1}{p{0.05\textwidth}|}{\centering RSF} &
		\multicolumn{1}{p{0.05\textwidth}}{\centering MAE} & \multicolumn{1}{p{0.07\textwidth}}{\centering RMSE} & 
          \multicolumn{1}{p{0.05\textwidth}}{\centering RSF} \\
\hline
\multirow{3}{*}  &
\multicolumn{1}{c|}{FairTP-DCRNN}  & $-1.20$\% &  $-1.95$\% & \multicolumn{1}{c|}{$+70.77$\%}  & $+25.65$\%  &  $+27.80$\%  &  $+42.16$\% \\
&\multicolumn{1}{c|} {FairTP-AGCRN}  &  $+4.91$\% &  $-3.46$\% &  \multicolumn{1}{c|}{$+64.98$\%}  &  $+31.21$\% & $+40.81$\% &  $+89.01$\% \\
 &  \multicolumn{1}{c|}{FairTP-GWNET} &  $+3.78$\% & $-3.26$\% &  \multicolumn{1}{c|}{$+58.77$\%}   & $+14.24$\% & $+13.05$\% &  $+49.30$\%  \\
  &  \multicolumn{1}{c|}{FairTP-ASTGCN} &  $-0.28$\% & $+0.27$\% &  \multicolumn{1}{c|}{$+89.97$\%}   & $+42.30$\% & $+53.68$\% &  $+88.96$\%  \\
  &  \multicolumn{1}{c|}{FairTP-DSTAGNN} &  $+13.03$\% & $+7.86$\% &  \multicolumn{1}{c|}{$+65.00$\%}   & $+12.29$\% & $+19.04$\% &  $+70.17$\%  \\
  &  \multicolumn{1}{c|}{FairTP-DGCRN} &  $+8.89$\% & $+3.15$\% &  \multicolumn{1}{c|}{$+30.06$\%}   & $+50.28$\% & $+54.49$\% &  $+61.80$\%  \\
  
  & \multicolumn{1}{c|}{FairTP-D2STGNN} & $-1.14$\% & $-1.98$\% &  \multicolumn{1}{c|}{$+27.84$\%}   & $+32.48$\%  &  $+37.83$\% &  $+73.70$\%   \\ \hline
  \end{tabular}
       \end{adjustbox}
         \end{flushleft}
\label{trade_off}
\end{table}


\end{document}